\documentclass[letterpaper, 10 pt, conference]{ieeeconf}  % Comment this line out if you need a4paper

\IEEEoverridecommandlockouts                              % This command is only needed if 
\usepackage{cite}
\usepackage{amsmath,amssymb,amsfonts,empheq}
\usepackage{mathtools}
\usepackage{textcomp}
\usepackage{url}
\usepackage{caption}
\usepackage{subcaption}
\usepackage{graphics}
\usepackage{epsfig} % for postscript graphics files
\usepackage[framemethod=tikz]{mdframed}
\usepackage{tablefootnote} % for table footnotes
\usepackage{algorithm}
\usepackage{algpseudocode}
\usepackage{titlecaps}
\usepackage{bbm}
\usepackage{breqn}
\usepackage{algorithm}
\usepackage{algpseudocode}
\usepackage{multicol}
\usepackage{multirow}
\usepackage{array}
\usepackage{comment}
\usepackage{booktabs}
\newcolumntype{N}{>{\centering\arraybackslash}m{.3in}}

\makeatletter
\let\NAT@parse\undefined
\makeatother
\usepackage{hyperref}

\newcommand{\sgn}{\operatorname{sgn}}

\title{\LARGE \bf
Re4MPC: Reactive Nonlinear MPC for Multi-model \\ Motion Planning via Deep Reinforcement Learning
}

\author{Ne\c{s}et \"{U}nver Akmandor$^{1*}$, Sarvesh Prajapati$^{2 \ddagger}$, Mark Zolotas$^{3 \ddagger}$, and Ta\c{s}k{\i}n Pad{\i}r$^{4}$% <-this % stops a space
\thanks{\tt\small $^{*}$akmandor.n@northeastern.edu}%
\thanks{$^{1}$Ne\c{s}et \"{U}nver Akmandor is currently at Motional AD Inc., Boston, MA, USA. This paper describes work performed at Northeastern University and is not associated with Motional AD Inc.}%, Boston, MA, 02115, USA}%
\thanks{$\ddagger$ Equally contributed authors.}%
\thanks{$^{2}$Institute for Experiential Robotics, Northeastern University, Boston, Massachusetts, USA.}
\thanks{$^{3}$Mark Zolotas is currently at Toyota Research Institute (TRI), Cambridge, MA, USA. This paper describes work performed at Northeastern University and is not associated with TRI.}
\thanks{$^{4}$Ta\c{s}k{\i}n Pad{\i}r holds concurrent appointments as a Professor of Electrical and Computer Engineering at Northeastern University and as an Amazon Scholar. This paper describes work performed at Northeastern University and is not associated with Amazon.}
}
\begin{document}

\maketitle
\thispagestyle{empty}
\pagestyle{empty}

%%%%%%%%%%%%%%%%%%%%%%%%%%%%%%%%%%%%%%%%%%%%%%%%%%%%%%%%%%%%%%%%%%%%%%%%%%%%%%%%
\begin{abstract}

Traditional motion planning methods for robots with many degrees-of-freedom, such as mobile manipulators, are often computationally prohibitive for real-world settings. In this paper, we propose a novel multi-model motion planning pipeline, termed Re4MPC, which computes trajectories using Nonlinear Model Predictive Control (NMPC). Re4MPC generates trajectories in a computationally efficient manner by reactively selecting the model, cost, and constraints of the NMPC problem depending on the complexity of the task and robot state. The policy for this reactive decision-making is learned via a Deep Reinforcement Learning (DRL) framework. We introduce a mathematical formulation to integrate NMPC into this DRL framework. To validate our methodology and design choices, we evaluate DRL training and test outcomes in a physics-based simulation involving a mobile manipulator. Experimental results demonstrate that Re4MPC is more computationally efficient and achieves higher success rates in reaching end-effector goals than the NMPC baseline, which computes whole-body trajectories without our learning mechanism.

%The dataset, the code, and further details including additional visualizations are available at \href{https://lhy.xyz/stereovoxelnet/}{https://lhy.xyz/stereovoxelnet/}.

\end{abstract}

%%%%%%%%%%%%%%%%%%%%%%%%%%%%%%%%%%%%%%%%%%%%%%%%%%%%%%%%%%%%%%%%%%%%%%%%%%%%%%%%
\section{Introduction}
\label{sec:introduction}

% \subsection{Motivation}
% \label{sec:motivation}

Motion planning in dynamic environments has been a significant challenge for robotics research in recent decades~\cite{mohanan2018survey}. This is primarily because limited computational resources compromise the quality of trajectory planning, as the robot's planner strives to fulfill task requirements while accounting for dynamic environmental changes. While advances in computation through modern GPUs and parallel processing have accelerated motion planning in robotic systems, substantial challenges remain for robots with high-dimensional configuration spaces, such as mobile manipulators and humanoids. The inherent complexity of the kinematics and dynamics of these robots makes real-time processing a significant challenge, thus affecting the robot's ability to respond rapidly to environmental changes~\cite{thakar2023survey}.

Moreover, the trade-off between using simplified and complex robot models for motion planning introduces further challenges. Simplified models are generally suitable for flat and predictable environments, but are inadequate for more unstructured settings, such as uneven terrains. By contrast, complex models require increased computational power and more sophisticated algorithms. These requirements for complex models present challenges to the scalability and efficiency of motion planning in diverse task scenarios~\cite{styler2017plan,norby2024adaptive}.

\begin{figure}[t]
\centering
\includegraphics[width=0.92\columnwidth]{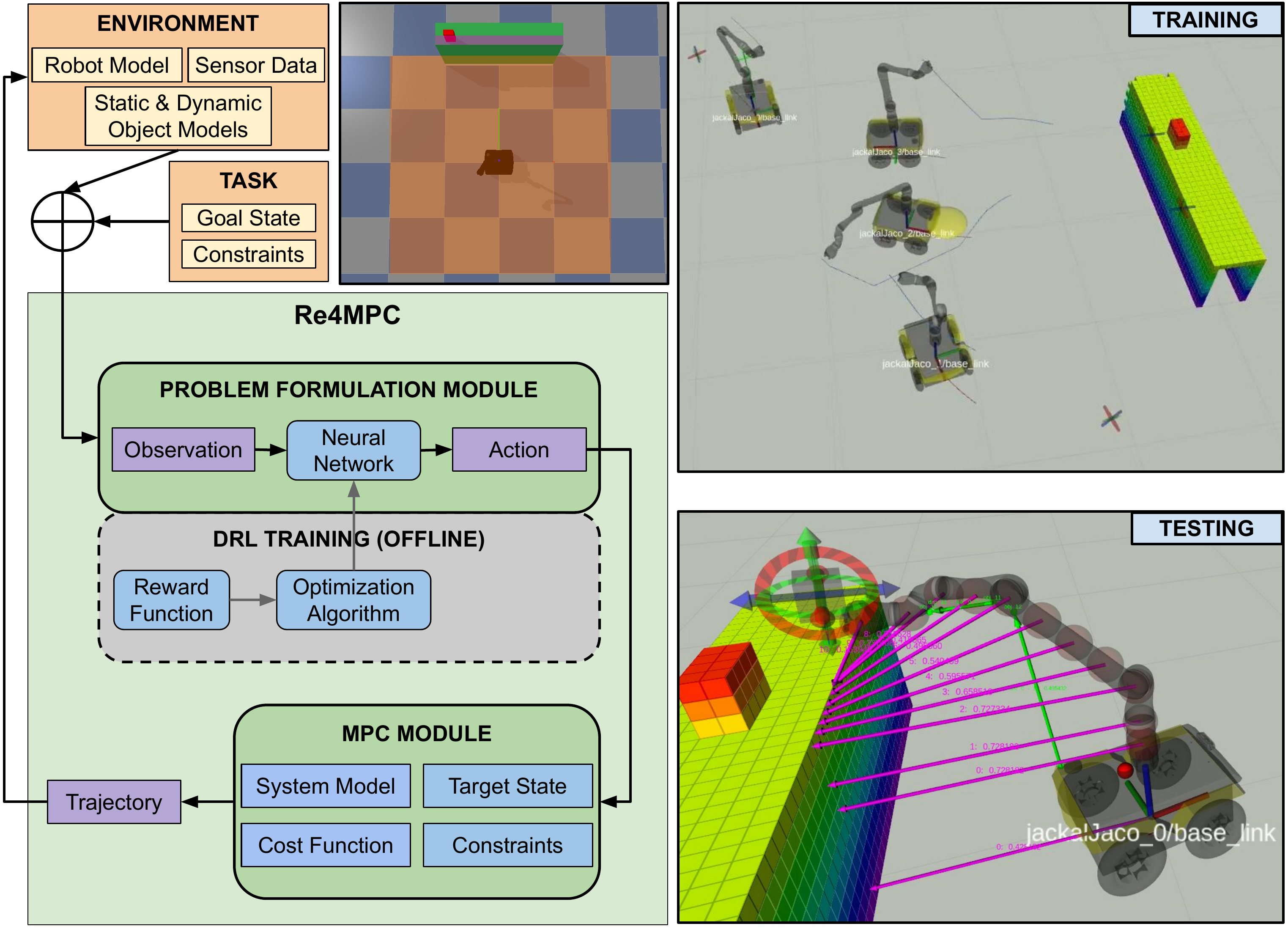}
\caption{The Re4MPC pipeline consists of two main components: 1) the ``Problem Formulation Module'' where the robot model, target state, cost function and constraints are learned by DRL; 2) the ``MPC Module'', which generates command trajectories of a robot model for some time horizon.}
\label{fig:re4mpc_framework}
\vspace{-3.2mm}
\end{figure}

Model Predictive Control (MPC) is widely recognized in robotics as an effective tool for motion planning due to its ability to predict future states and adjust in real-time~\cite{quirynen2024real}. However, the effectiveness of MPC largely depends on how well it can be adapted to various tasks and environments. For example, strategies that work well for motion planning in open and unobstructed spaces may not be suitable in crowded or constrained areas. Hence, the MPC formulation--especially the cost function and constraints--must be modified in order to suit each specific scenario~\cite{adajania2022multi}.

To address these issues, we present \textbf{Reactive Multi-model MPC (Re4MPC)}, a methodology designed to enhance the computational efficiency of motion planning, particularly for robotic systems with high-dimensional configuration spaces. Our approach leverages both complete and subsystem dynamic models, available in varying degrees of complexity. As a result, the main objective of our work is to achieve higher success rates while reducing online computation costs for tasks that demand rapid response times in dynamic environments. The core contributions of this paper are:
\begin{itemize}
\item A novel motion planning pipeline, Re4MPC, which efficiently computes trajectories using Nonlinear MPC (NMPC) settings that are reactively adapted according to the robot state and task. The policy for adjusting these settings is learned via a Deep Reinforcement Learning (DRL) framework, for which we provide a formulation.
\item An evaluation of Re4MPC and our design choices, including findings and ablation studies from experiments conducted in a physics-based simulator.
\item An open-source implementation of Re4MPC and code to reproduce our experiments: \url{https://github.com/RIVeR-Lab/mobiman}.
\end{itemize}

\section{Related Work}
\label{sec:related}

\subsection{Classical Methods} \label{sec:re4mpc:related_classical}

For robots with many degrees-of-freedom (DOF), such as mobile manipulators and humanoids, there are several classical motion planning methods that are computationally feasible in the presence of dynamic objects~\cite{mohanan2018survey}. While node-based and sampling-based methods face scalability issues in high-DOF robotic systems due to the ``curse of dimensionality'', a select few studies~\cite{saxena2021manipulation,carius2022constrained} have demonstrated practical results. Reactive methods~\cite{haviland2022holistic,burgess2023architecture} have proven highly effective in responding to dynamic changes in the environment and executing tasks that require online adaptation. Nonetheless, their task-dependent formulation is often inefficient when generalizing across different tasks. 

Optimization-based methods are another popular approach to motion planning due to their mathematical guarantees~\cite{spahn2021coupled}. Compared to reactive approaches, these methods provide a scalable solution that is suitable for any robotic system~\cite{dong2023review}. MPC-based methods also exhibit faster runtime performance via gradient-based optimization than sampling-based alternatives. Specifically,~\cite{pankert2020perceptive,sleiman2021unified} utilized a Differential Dynamic Programming (DDP) based NMPC approach, known as Sequential Linear Quadratic (SLQ)~\cite{sideris2005efficient,sleiman2021constraint}. These works demonstrated SLQ's versatility in managing locomotion of legged robots and whole-body control of various mobile manipulator types (legged or wheeled), while accounting for factors, such as collision avoidance, mechanical stability, and task-specific constraints.

\subsection{DRL Methods} \label{sec:re4mpc:related_fusion}

Model-free and end-to-end DRL approaches have shown impressive results for motion planning~\cite{dong2023review}. Despite these results, convergence to a desired policy can prove challenging and lead to lengthy training times, as state and action space dimensionality expands for complex task settings. Prior work~\cite{wang2020learning,iriondo2023learning} has trained end-to-end policies for whole-body control of mobile manipulators by crafting task-specific reward functions to guide the agent toward the goal. To minimize the training duration for long-horizon and interactive mobile manipulation tasks,~\cite{li2020hrl4in} introduced a hierarchical architecture that enabled the agent to learn both high-level policies, such as defining parts of the system model and their corresponding sub-goals, as well as low-level control policies. Similarly, our DRL framework learns high-level policies; however, unlike the aforementioned methods, our system uses an NMPC method for low-level control.

Motion planning methodologies that integrate classical approaches with DRL have been widely explored to enhance both the efficiency of policy training and adaptability across diverse tasks~\cite{bellegarda2020online,yamada2021motion,xia2021relmogen,grandesso2023cacto}. A common theme among these studies is using DRL to generate high-level goals or adjust parameters within a motion planner, rather than directly outputting low-level control commands from a neural network. This hybrid strategy has been shown to accelerate convergence to an optimal policy compared to end-to-end learning of control actions~\cite{yamada2021motion,xia2021relmogen}. Our primary contribution is to extend the applicability of such hybrid strategies by enabling multi-model selection in NMPC for mobile manipulation.

Several prior works have leveraged MPC in conjunction with DRL, but they have typically focused on either mobile bases or robotic arms separately. For example, some studies~\cite{yang2020data,feher2020hierarchical} used MPC to compute local trajectories for mobile robots, whereas~\cite{xia2021relmogen} applied a sampling-based algorithm to plan trajectories for either the base or arm of a mobile manipulator by employing a hierarchical training strategy. However, these approaches do not provide a unified whole-body motion planning framework, limiting their ability to optimize the entire robot's motion in a single process.

Our proposed method, Re4MPC, establishes multi-model selection within the NMPC formulation to generalize how DRL is integrated with NMPC. This multi-model selection mechanism allows full-body or partial-body trajectories to be efficiently computed within a single framework. Among studies combining MPC with DRL, our work is distinctive for its emphasis on mobile manipulation and multi-model adaptability. Our implementation also contributes an open-source framework that integrates this multi-model feature into the well-documented NMPC library: OCS2~\cite{OCS2}. Stable-Baselines3~\cite{stable-baselines3} is also leveraged for DRL training and PyBullet~\cite{coumans2021} is used for physics-based simulation and evaluation.

%===============================================================================
\section{Methodology}
\label{sec:methodology}

The Re4MPC pipeline, shown in Fig.~\ref{fig:re4mpc_framework}, consists of two main components: 1) the policy, learned by DRL, which formulates the NMPC problem by determining the robot model, target, and constraints; 2) NMPC to generate command trajectories of a robot model for some time horizon.

\subsection{Multi-model Nonlinear Model Predictive Control}

\subsubsection{Kinematic Robot Models} \label{sec:kinematic}

A discrete set of robot models $F$, which may consist of either kinematic or dynamic equations of motion, is used to compute their respective future trajectories. As a proof of concept, we consider three different models of a custom-designed mobile manipulator platform, where we use Clearpath's Jackal robot as our non-holonomic mobile base, with a 6-DOF robotic arm, Kinova's Gen-1 Jaco, mounted on its chassis.

Therefore, the discrete set of robot models in our work is $F=\{f_{b},f_{a},f_{wb}\}$, for base, arm, and whole-body, respectively. We define the mobile base state as $s_b = [x_b, y_b, \phi_b]$, which is comprised of 2D coordinate positions $x_b$ and $y_b$, as well as yaw angle $\phi_b$, each relative to a global coordinate frame. Control input $u_b = [v_{x_b}, \omega_{\phi_b}]$ consists of translational velocity $v_{x_b}$ and angular velocity $\omega_{\phi_b}$. A non-slip skid-steer robot model~\cite{rabiee2019friction} represents the kinematics $\dot{s}_b = f_b(s_b, u_b)$ of the non-holonomic mobile base. For the arm model, we use $N_a$ joint angles $\theta_1,\ldots,\theta_{N_a}$ to define its state $s_a$. Using velocity control $u_a=[u_{\theta_{1}},..,u_{\theta_{N_a}}]$, the kinematic equation of the arm becomes $\dot{s}_a=f_a(s_a, u_a)=[\dot{\theta}_{1},\dot{\theta}_{2},\ldots,\dot{\theta}_{N_a}]^T$. Finally, the kinematic equation $\dot{s}_{wb} = f_{wb}(s_{wb},u_{wb})=[\dot{s}_b, \dot{s}_a]$ for whole-body motion is a vector concatenation of both the mobile base and robot arm kinematics, since the system equations are decoupled. 

\subsubsection{Multi-model SLQ}

For each model $f_m(s_m, u_m) \in F$, where $m = 1, \ldots, N_m$ and $N_m$ is the number of robot kinematic models, we formulate our NMPC problem~\cite{sideris2005efficient,sleiman2021constraint} as:
\begin{equation} \label{eq:mpc_multi}
\begin{aligned}
    \min_{u_m} \quad & J_{T_m} + \int_{t=t_0}^{t_0+T_h} J_{I_m}(t) \,dt\\
    \textrm{s.t.} \quad & \dot{s}_m = f_m(s_m, u_m) \\
      & g_{mi}(s_m, u_m) = 0, \, \forall i     \\
      & h_{mj}(s_m, u_m) \geq 0, \, \forall j.
\end{aligned}
\end{equation}
Here, the aim is to minimize the sum of terminal $J_{T_m}$ and intermediate costs $J_{I_m}$ with respect to the control input $u_m$. This minimization is subject to constraints from the robot model $f_m$, as well as sets of equality $G_m \in {g_{mi}(s_m, u_m)}$ and inequality $H_m \in {h_{mj}(s_m, u_m)}$ constraints. Subscripts $i\in [0, N_{ch_m}]$ and $j\in [0, N_{cg_m}]$ are indices of the equality and inequality constraints for model $m$. The total number of constraints in the set $C={H_0 \cup \ldots \cup H_m \cup G_0 \cup \ldots \cup G_m}$ is thus $N_c=\sum_{m=1}^{N_m} (N_{ch_m} + N_{cg_m})$.

To compute trajectories for a specific robot model, we use an NMPC method, known as SLQ~\cite{sideris2005efficient,sleiman2021constraint}. This algorithm operates through a series of computational steps that include forward integrating the system dynamics using nominal control laws to generate trajectories. SLQ then linearizes the system dynamics around these nominal trajectories to formulate an approximate Linear-Quadratic (LQ) problem. Additionally, a line-search mechanism is used to iteratively adjust control updates, ensuring that each step results in an improvement over the previous cost. Based on the model index $m$, the SLQ algorithm takes as inputs the dynamic model $f=f_m$, terminal $J_T=J_{T_m}$ and intermediate $J_I=J_{I_m}$ costs, constraint set $C$, current state $s(t_0)=s_m(t_0)$, target state $\hat{s}(T_h)=\hat{s}_m(T_h)$ at the trajectory's time horizon $T_h$, and the maximum MPC iterations $N_{mpc}$. SLQ returns the most recent control policy $u$ and computation time $\Delta t_p$.  

Our cost function is designed to minimize both control input and pose error between the end-effector and target. As DDP cannot handle constraints, we employ the Relaxed Barrier Function (RBF) methodology~\cite{feller2016relaxed}. In our experiments, system constraints only consider joint position and velocity limits to simplify analysis. However, our approach can naturally incorporate other task-specific constraints, such as self-collision or external collision constraints. Additionally, since collision constraints and mechanical stability are accounted for by our reward function, our NMPC formulation indirectly considers these factors during DRL training.

\subsection{DRL Framework to Determine NMPC Settings}

Our Re4MPC motion planning strategy uses DRL to train a neural network that outputs different possible NMPC settings, as described in Eq.~\ref{eq:mpc_multi} and summarized in Algorithm~\ref{algo:re4mpc}. Each iteration $n$ of the algorithm begins by selecting an action $a \in A$ based on the current observation $o \in O$. The chosen action is decoded to provide essential information to the NMPC solver, including: 1) the index $m$ of the robot model $f_m$; 2) terminal $J_{T_m}$ and intermediate $J_{I_m}$ costs; 3) an updated constraint set $C_m$; and 4) the target state $\hat{s}_m$ at the end of the time horizon $T_h$. This information is utilized for $T_{action}$ seconds to iteratively compute and execute whole-body control commands. During this interval, the model state $s_m$ is derived from recent observations $o$ and fed into an NMPC algorithm, such as SLQ, along with the other aforementioned inputs. To train the policy network, data tuples consisting of $o(t)$, $a(t+t_p)$, $r(t+t_p)$, and $o(t+t_p)$ are collected and added to the current sample set $D$.

\begin{algorithm}[t] 
\caption{Reactive Multi-model NMPC (Re4MPC)} \label{algo:re4mpc}
\renewcommand{\algorithmicrequire}{\textbf{Input:}}
\renewcommand{\algorithmicensure}{\textbf{Output:}}
\begin{algorithmic}[1] 
\Require $env$, $\textit{NMPC}$, $J_T$, $J_I$, $C$, $T_{action}$, $N_{rl}$, $N_{mpc}$, $T_h$, $D$
\Ensure $D$
\For {$n=1,2,\ldots, N_{rl}$}
    \State $a(t) \gets \pi(o(t))$
    \State $m \gets getModelMode(a(t))$
    \State $J_{T_m}, J_{I_m} \gets getModelCost(a(t), J_T, J_I)$
    \State $C_m \gets getModelConstraints(a(t), C)$
    \State $\hat{s}_m(t+T_h) \gets getModelTarget(a(t))$
    \State $t_p \gets 0$
    \While{$t_p < T_{action}$}
        \State $s_m(t+t_p) \gets getModelState(o(t+t_p))$
        \State $u_m, \Delta t_p \gets \textit{NMPC}(f_m,J_{T_m},J_{I_m},C_m,s_m(t+t_p),\hat{s}_m(t+t_p+T_h),T_h,N_{mpc})$
        \State $a_{wb}(t+t_p) \gets mapWholeBodyControl(u_m)$
        \State $o(t+t_p+\Delta t_p), r(t+t_p+\Delta t_p) \gets env.step(a_{wb}(t+t_p))$
        \State $t_p \gets t_p+\Delta t_p$
    \EndWhile
    \State $D \gets D \cup \{[o(t), a(t+t_p), r(t+t_p), o(t+t_p)]\}$
\EndFor
\end{algorithmic}
\end{algorithm}

\subsubsection{Observation Space}
\label{sec:observation}

The decision-making of Re4MPC is modeled as a Partially Observable Markov Decision Process (POMDP)~\cite{kaelbling1998planning}, which assumes the agent's global state (e.g., the mobile base position) is unobservable. Hence, our observation space $\Omega$ only includes information obtainable via sensor data and defined as $o = [o_{goal}, o_{occ}, o_{self}, o_{v_b}, o_{\phi_b}, o_{theta}, o_{\omega}] \in \Omega$.  

We define a goal observation as $o_{goal} = {}^Rp_g$ based on its position ${}^Rp_{g}$ relative to the robot's coordinate frame $R$. For each occupancy object $k$, we encapsulate 3D position and volumetric information as ${}^Rp_{occ}=[{}^Rp_{occ_1},\ldots,{}^Rp_{occ_{N_{occ}}}] \in \mathbb{R}^{6\times N_{occ}}$, where ${}^Rp_{occ_k} = [{}^Rx_{occ_k}, {}^Ry_{occ_k}, {}^Rz_{occ_k}, w_{occ_k}, l_{occ_k}, h_{occ_k}]$ and $N_{occ}$ is the total number of occupancy objects. For the self-collision observation, we compute the minimum distance between $N_{self}$ link pairs, with each observation $o_{self} \in \mathbb{R}^{N_{self}}$. Additionally, we incorporate lateral $o_{v_b}$ and angular $o_{\phi_b}$ velocities of the mobile base, joint positions $o_{\theta}$, and angular velocities $o_{\omega}$ of the arm joints into the observation space. 

Please note that our observation space was designed to balance computational efficiency with task-specific effectiveness. We acknowledge that more generalizable observation space designs, such as raw sensory inputs from vision or point cloud data, or more sophisticated representations as in~\cite{akmandor2022deep}, could enhance adaptability to novel environments. However, incorporating such representations would increase training times and may require more complex policy architectures to extract meaningful features. As we are primarily focused on establishing and evaluating the Re4MPC framework for training and selecting models of varying complexity, we elected a structured and simple observation space to ensure feasibility within the given constraints. Nevertheless, the core framework does not inherently depend on any specific observations and could be extended.

\subsubsection{Action Space}
\label{sec:action}

Similar to~\cite{yamada2021motion,xia2021relmogen}, we employ a parameterized action space instead of learning direct control commands. At each RL step, the selected action is decoded to formulate the NMPC problem specified in Eq.~\ref{eq:mpc_multi}. This action space representation significantly enhances computational efficiency and increases the task success rate by allowing robot models and constraints to be adaptively configured during trajectory computation. As the NMPC solver may fail to compute a stable trajectory for a given goal~\cite{grandesso2023cacto}, we train the RL agent to decompose a goal into achievable sub-goals (targets) using the latest observations. To accommodate various RL algorithms and their implementations, we developed two types of action spaces: a continuous space $A_c$ and a discrete space $A_d$. Both types include sub-action spaces that determine the robot model $a_{model}$, regulate constraints $a_{constraint}$, and select the next target $a_{target}$.  

The sub-action spaces for $A_c$ are defined as continuous variables and concatenated into a 1D vector, such that $A_c = [a_{model}, a_{constraint}, a_{target}]$. The model index $m$ is determined using $N_m-1$ thresholds, which equidistantly divide the continuous space of $a_{model}$. For $a_{model} \in [0,1]$, $m=0$ if $a_{model} \leq 0.3$, $m=1$ if $0.3 < a_{model} \leq 0.6$, and $m=2$ if $a_{model} > 0.6$. Similarly, to decide whether a model constraint $h_{mji}$ or $g_{mj}$  is included in the problem formulation or not, each action's constraint subspace $a_{constraint} \in [0, 1]^{N_c}$ is checked against a threshold of $0.5$.

The first dimension of the target action sub-space determines the target's type $a_{type} \in [0,1]$, allowing the agent to decide whether the next end-effector target is a sub-goal pose, $a_{type} \leq 0.5$, or directly the goal pose, $a_{type} > 0.5$. The remaining sub-spaces are used to set a 3D target pose, such that $a_{target} = [a_{type}, a_{x}, a_{y}, a_{z}, a_{\psi}, a_{\vartheta}, a_{\phi}] \in \mathbb{R}^7$. To accommodate both target types, sub-goal and direct goal, $a_{x}, a_{y}, a_{z}$ are defined in $[-1, 1]$ and $a_{\psi}, a_{\vartheta}, a_{\phi}$ in $[-\pi, \pi)$. Depending on $a_{type}$, the target position values $a_{x}, a_{y}, a_{z}$ are mapped to a continuous range $[\alpha_{min_{k}}, \alpha_{max_{k}}]$, where $k \in \{x, y, z\}$. When $a_{type} \leq 0.5$, $\alpha_{min_{k}}$ and $\alpha_{max_{k}}$ are defined relative to the robot's coordinate frame. On the other hand, when $a_{type} > 0.5$, these range values are adjusted to account for localization errors with respect to the goal frame.  

In addition to $A_c$, we define a discrete action space $A_d$ to specify the model, constraints, and target pose. We discretize each of the aforementioned action sub-spaces and assign each combination an index.

\subsubsection{Reward Function}
\label{sec:reward}

To converge to a policy function that maximizes total discounted reward, the reward function must guide the agent toward the desired goal state. Similar to~\cite{wang2020learning,xia2021relmogen}, we design our reward function considering both terminal and intermediate states. Terminal rewards $R_{term}$, which can be either positive or negative constant values, are triggered when the robot either reaches the goal state $\tau_{success}$, goes out of bounds $\tau_{boundary}$, collides with itself or an external object $\tau_{collision}$, rolls over $\tau_{roll}$, or exceeds the maximum allowable time $\tau_{max\_step}$:
\begin{equation}
   \text{$R_{term}$}= 
\begin{cases}
   \tau_{success},&\text{if } d_{pos}(p_{ee}, p_g) + d_{ori}(q_{ee}, q_g) < \delta_{g}, \\
   \tau_{boundary},&\text{if } x_b < \delta_{x_{min}} \text{ or } x_b > \delta_{x_{max}} \text{ or} \\
   & \quad y_b < \delta_{y_{min}} \text{ or } y_b > \delta_{y_{max}} \text{ or} \\
   & \quad z_b < \delta_{z_{min}} \text{ or } z_b > \delta_{z_{max}} \\
   \tau_{collision},&\text{if } \exists j, k \quad d_{pos}(p_{cp_j}, p^*_{occ_k}) < \delta_{cd}, \\
   \tau_{roll},&\text{if } \psi_{b} > \delta_{\psi} \text{ or } \vartheta_{b} > \delta_{\vartheta}, \\
   \tau_{max\_step},&\text{if } n > N_{rl}
\end{cases}
\end{equation}
We utilize two distance metrics, Euclidean distance $d_{pos}$ and angular distance $d_{ori}$ between two quaternions, as described in~\cite{siciliano2009}, to compute the position and orientation errors between the end-effector pose and goal pose. If the sum of these position and orientation errors becomes less than a threshold, we assume the goal is reached.

During training, we define minimum and maximum volumetric boundaries in the robot's workspace to reduce exploration time. When the robot's position exceeds these ranges, we penalize the RL agent by terminating the episode.  

To determine external collisions, we sample a set of collision points between the actuated links of the robot. Each timestep, we compute distances between all collision points and their closest points on the $N_{occ}$ occupancy objects. We reset the training environment with a termination penalty if any of the computed distances is less than a collision distance threshold. We also account for collisions with the ground if any collision point's height is below some threshold. Self-collision is determined if the closest distance between any link pairs is less than a threshold.  

At each iteration $n$, the step reward $R_{step}$ is calculated as the weighted sum of multiple reward and penalty terms: $R_{model}$ penalizes computation of the active system model, $R_{goal}$ rewards the robot end-effector's closeness to goal, and $R_{target}$ evaluates the end-effector based on the target pose while also assessing mechanical stability given the arm's joint states. To discourage the robot from wandering about and to encourage the selection of efficient system models, we design $R_{model}$ as a constant value penalty function where the penalty values are proportional to the DOF of the chosen system model. To drive the learning agent toward the goal, we design $R_{goal}$ as the distance between the goal and the robot’s end-effector across consecutive action steps.

We motivate the agent to choose sub-targets over the goal pose directly by introducing a target reward $R_{target}$. In our $R_{target}$ design, there are distinct reward terms for \textit{arm-only} and \textit{other than arm-only} modes. This distinction is necessary because the quality of an \textit{arm-only} trajectory is associated with the mechanical balance of the robot, while trajectories with other modes are mainly guided by navigation-related heuristics. In the \textit{arm-only} mode, we compute the average center-of-mass (CoM) position difference $\Omega_{com}$ over the action horizon using an empirical CoM position $p_{com}^{*}$ where the robot remains stable. $R_{\text{target}}^{\text{arm}}$ is then determined by a positive constant term $\tau_{\text{target}}$, serving as a penalty if $\Omega_{\text{com}}$ exceeds a threshold value $\delta_{\text{com}}$ and as a reward otherwise, based on the output of the signum function:
\begin{align}
    \label{eq:reward_target_arm}
    &R_{target}^{arm} = \sgn{(\delta_{com} - \Omega_{com})} \tau_{target} \\
    &\text{where } \Omega_{com} = \frac{\sum_{t_p=0}^{N_{t_p}-1} d_{pos}(p_{com}(t_p), p_{com}^{*}(t_p))}{N_{t_p}}. \nonumber
\end{align}

For \textit{other than arm-only} modes, the target reward $R_{target}^{base}$ is determined based on whether the sub-target is reached. If the sub-target is not reached, we compute $R_{target+}^{base}$, which is similar to $R_{goal}$. Instead of using the goal pose, the agent is rewarded for approaching a sub-target position $p_{sub}$ and yaw orientation $\phi_{sub}$. A simple heuristic determines the sub-goal pose by interpolating along the line connecting the 2D positions of the robot $p_b$ and the goal $p_g$, scaled by a user-defined parameter $\alpha_{sub}$. To compute the sub-goal's yaw angle $\phi_{sub}$, we use a stable implementation of the $\arctan$ function, $atan2$. Once $p_{sub}$ is established, the position difference $\Delta p$ between the robot and the sub-target is normalized to the range $[0,1]$. Similarly, the yaw difference $\Delta \phi$ is normalized to $[0,1]$. Since the robot can rotate in both directions, the minimum yaw difference with respect to either $0$ or $\pi$ radians is considered. The reward $R_{\text{target+}}^{\text{base}}$ is then calculated using an exponential function, with the equally weighted sum of $\Delta p$ and $\Delta \phi$ as input and a negative scalar $\gamma$. Similar to $R_{target}^{arm}$, the magnitude of the reward is determined by the positive reward term $\tau_{\text{target}}$:
\begin{align}
    \label{eq:reward_target_base}
    &R_{target+}^{base} = \tau_{target} e^{0.5 \gamma (\Delta p + \Delta \phi)} \\
    &\text{where}\hspace{0.5em} \Delta p = \frac{d_{pos}(p_{b}^{n-1}, p_{sub}) - d_{pos}(p_{b}^{n}, p_{sub})}{d_{pos}(p_{b}^{n-1}, p_{sub})}, \nonumber \\
    &\quad\quad\quad \Delta \phi = \frac{\min(\pi - (\phi_{sub} - \phi_b^{n}), (\phi_{sub} - \phi_b^{n}))}{\pi}, \nonumber \\
    &\quad\quad\quad p_{sub} = p_b + \alpha_{sub} (p_g - p_b), \nonumber \\
    &\quad\quad\quad \phi_{sub} = \text{atan2}(y_g - y_b, x_g - x_b). \nonumber
\end{align}

Although moving toward effective sub-targets improves motion planning performance, prematurely reaching them within the action horizon negatively impacts control flow. To prevent unnecessary interruptions and enhance the smoothness of consecutive actions, a penalization term $R_{target-}^{base}$ is introduced when the base target is reached:
\begin{align}
\label{eq:reward_target_penalty}
    &R_{target-}^{base} = -\tau_{target} \frac{e^{-\gamma \Delta p}}{e^{\gamma}}
\end{align}

% Overall, the step target reward is calculated as:
% \begin{equation} 
%    \label{eq:reward_target}
%    \text{$R_{target}$}= 
% \begin{cases}
%    R_{target\_arm},& \text{if } f_m = f_a \\
%    R_{target\_penalty},& \text{else if } d_{pos}(p_{ee}, p_{sub}) \\
%    & \text{           } + d_{ori}(q_{ee}, q_{sub}) < \delta_{g} \\
%    R_{target\_base},& \text{else }
% \end{cases}
% \end{equation}

%===============================================================================

\section{Results and Analysis}
\label{sec:results}

Re4MPC is built atop ROS with the OCS2~\cite{OCS2} library used to implement our multi-model NMPC approach. The iGibson $2.0$ library~\cite{igibson} and its PyBullet~\cite{coumans2021} physics engine are used to simulate our experiments. We designed a fully connected neural network with two hidden layers (400x300) and ReLU activations to model our policy function.

Our training and testing environments are designed to demonstrate that the trained approach effectively adapts its MPC formulation by switching models and assigning feasible sub-goals based on the current state. While policy robustness and generalization to unseen environments are beyond this paper’s scope, these could be improved through training to convergence across diverse environments and tasks, as in~\cite{li2020hrl4in,yamada2021motion,xia2021relmogen}. For our training and experiments, we use a simple simulation with a stationary conveyor belt and a box to be picked up (Fig.~\ref{fig:re4mpc_framework}). Although the box position remains fixed during training, our reactive planner can handle dynamic objects when the environment is set accordingly, which we leave as future work.

\subsection{DRL Training}

At the start of each episode, the robot's base pose is randomly initialized within a designated area on the conveyor belt. While we set the end-effector (grasp) goal to a fixed orientation, its position is randomized on the conveyor belt’s surface. To minimize the required learning time, without aiming to enhance robustness, we do not randomize the initial joint positions of the arm or conveyor belt’s pose.  

We train the policy network in Re4MPC using three different RL algorithms: PPO~\cite{schulman2017proximal}, SAC~\cite{haarnoja2018soft}, and DQN~\cite{mnih2015human}. Continuous action space representations are used for PPO and SAC, while a discrete action space is set for DQN. To analyze the effect of the target reward, we also train policies (\textit{re4mpc-PPO-w/o$R_{T}$}, \textit{re4mpc-SAC-w/o$R_{T}$}, \textit{re4mpc-DQN-w/o$R_{T}$}) without including  $R_{target}$ in their reward calculation. As a baseline, we use an OCS2~\cite{OCS2} implementation of SLQ, \textit{ocs2wb}, to generate whole-body trajectories, where the goal pose is the MPC target at each iteration.

\begin{figure}[t]
\centering
\includegraphics[width=0.78\columnwidth]{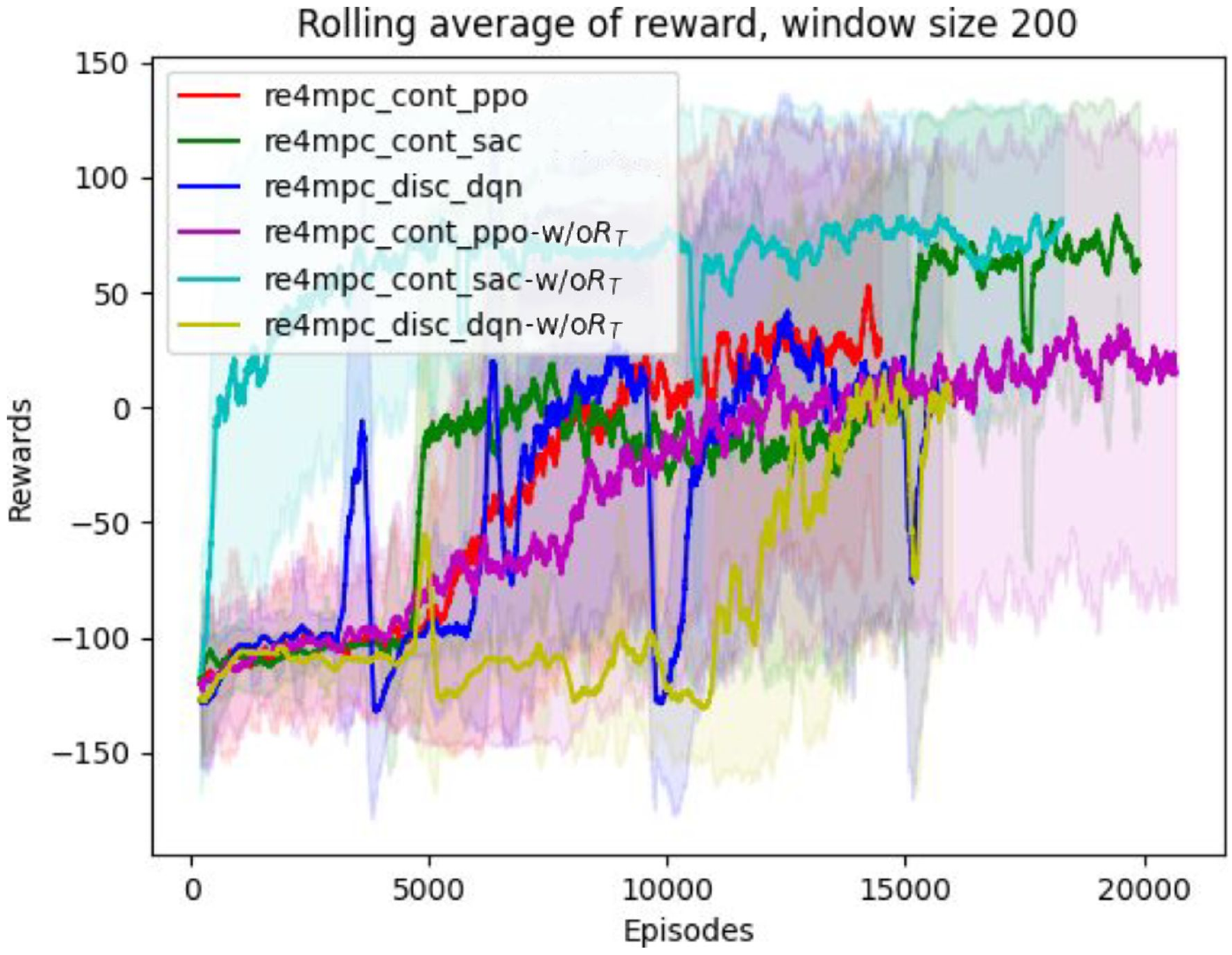}
\caption{Training curves for Re4MPC when trained using three different RL algorithms: PPO, SAC, and DQN. \textit{re4mpc-PPO} and \textit{re4mpc-SAC} use a continuous action space, while the action space of \textit{re4mpc-DQN} is discrete. The \textit{w/o$R_{T}$} methods are trained without the target reward $R_{target}$.}
\label{fig:reward_episodic_v0}
\vspace{-3.5mm}
\end{figure}

Fig.~\ref{fig:reward_episodic_v0} illustrates the training curves for these different DRL variants as a function of cumulative reward. After $15k$ episodes, all variants of Re4MPC begin to receive positive reward values, which are averaged over a window size of $200$ episodes. Both the SAC versions with \textit{re4mpc-SAC} and without the target reward \textit{re4mpc-SAC-w/o$R_{T}$} obtain the highest rewards, while the latter has the fastest convergence.

\subsection{Experimental Evaluation}

To evaluate the performance of our trained models, we initialize the robot and goal in $108$ different configurations within the same environment shown in Fig.~\ref{fig:re4mpc_framework}. These configurations are generated by equidistantly sampling the robot's sub-workspace along the $x$ and $y$ axes, the goal position's range on the $x$-axis, and the robot's yaw angle from $-\pi$ to $\pi$. We average results over five runs of each configuration.  

\begin{figure}[t]
\centering
\includegraphics[width=0.78\columnwidth]{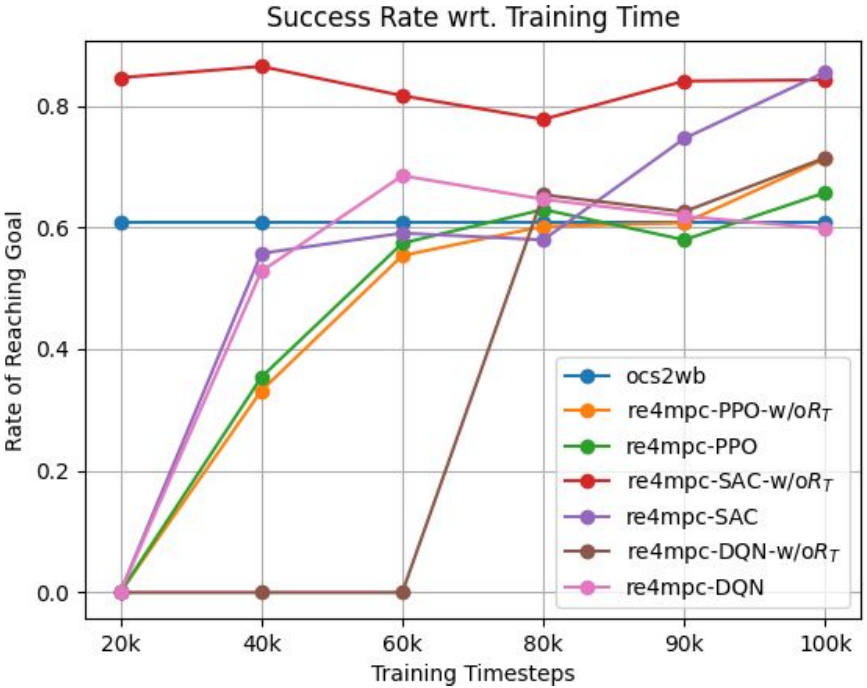}
\caption{Success rate per training timestep for the NMPC baseline and all Re4MPC variants.}
\label{fig:testing_result_arena}
\vspace{-3.5mm}
\end{figure}

The results depicted in Fig.~\ref{fig:testing_result_arena} show success rates for our baseline method and policy rollouts for all Re4MPC variants.  Here, \textit{re4mpc-SAC-w/o$R_{T}$} demonstrates the best performance, surpassing the baseline \textit{ocs2wb} by approximately $20$\% in success rate as early as $20k$ timesteps into training. After $100k$ training timesteps, all variants achieve higher success rates with fewer failures than the baseline.

To further assess the performance of trained models, we include the percentage of failure modes in Table~\ref{tab:testing_results}, such as out-of-boundary navigation, rollovers, self or external, and collisions. As expected, the baseline NMPC method prevents the robot from going out-of-boundary or exceeding the maximum timestep since trajectories are generated to reach the goal quickly. The main sources of failure for this method are collisions and rollovers. We highlight that rollovers are associated with its aggressive approach, as control magnitudes for both the arm and base are directly proportional to the distance between the robot's initial configuration and the goal pose. This often results in large momenta from arm motion, leading to instability in the robot's posture.  

One could argue that these rollovers are preventable by implementing additional constraints, such as increasing penalties for joint velocities or enhancing mechanical stability, as suggested in~\cite{pankert2020perceptive}. However, adding more constraints to the NMPC formulation complicates the optimization problem. For example, adding these constraints into the cost function without careful tuning of their weights might lead to unstable trajectories due to issues with local minima. On the other hand, the results shown Table~\ref{tab:testing_results} illustrate that Re4MPC significantly lessens rollover occurrence by switching to a reduced-order model and setting feasible sub-goals. This approach is effective without compromising joint velocities or introducing additional constraints. 

\begin{table}[t]
\caption{Test results on success rates and failure modes after $100k$ training timesteps, as well as number of trajectory computation calls and average computation time $\Delta t_p$ per selected model: base ($f_{b}$), arm ($f_{a}$), or whole-body ($f_{wb}$).}
\footnotesize
\centering
\setlength\tabcolsep{2 pt} % Essential for spacing
\begin{tabular}{c|c|c c|c c|c c}
\toprule
\textbf{Metric} & \textbf{ocs2wb} &  \textbf{PPO} & \textbf{w/o$R_T$} & \textbf{SAC} & \textbf{w/o$R_T$} & \textbf{DQN} & \textbf{w/o$R_T$} \\ 
\midrule
Success & 61 & 66 & 71 & \textbf{86} & 84 & 71 & 60 \\ \hline
Rollovers & 18 & 2 & 2 & \textbf{0} & 1 & 2 & 0 \\ \hline
Collisions & 21 & 29 & 24 & \textbf{11} & 12 & 24 & 35 \\ \hline
Out-of-boundary & \textbf{0} & 3 & 3 & 3 & 3 & 2 & 5 \\ \hline \hline
\# of $\Delta t_p$ if $f_{b}$ & 0 & \textbf{30201} & 20013 & 17696 & 21789 & 36244 & 24253 \\ \hline
Mean (\& std-dev) & 0 & \textbf{1.42} & 1.91 & 2.03 & 1.53 & 1.84 & 1.60 \\ 
$\Delta t_p$ if $f_{b} [ms]$ &  & \textbf{(0.05)} & (0.02) & (0.04) & (0.06) & (0.10) & (0.02) \\ \hline \hline
\# of $\Delta t_p$ if $f_{a}$ & 0 & 926 & 1480 & 381 & \textbf{1245} & 0 & 5387 \\ \hline 
Mean (\& std-dev) & 0 & 3.87 & 4.25 & 4.02 & \textbf{2.58} & 0 & 2.89 \\
$\Delta t_p$ if $f_{a} [ms]$ &  & (0.06) & (0.11) & (0.06) & \textbf{(0.13)} &  & (0.09) \\ \hline \hline
\# of $\Delta t_p$ if $f_{wb}$ & 24178 & \textbf{11093} & 14369 & 22588 & 20868 & 5700 & 15711 \\ \hline
Mean (\& std-dev) & 5.46 & \textbf{3.44} & 4.56 & 4.06 & 5.63 & 4.20 & 4.58 \\
$\Delta t_p$ if $f_{wb} [ms]$ & (0.49) & \textbf{(0.14)} & (0.28) & (0.15) & (0.25) & (0.03) & (0.34) \\
\bottomrule
\end{tabular}
\label{tab:testing_results}
\vspace{-4.0mm}
\end{table}

To quantify the computational efficiency gained by our approach, Table~\ref{tab:testing_results} presents the mean and standard deviation of computation time for each robot model selected during testing. It can be observed that employing only the base model to compute the trajectory reduces the computation time by more than half across all Re4MPC variants. It is important to note that MPC computation time is highly dependent on the target pose relative to the robot's initial configuration. Specifically, \textit{ocs2wb} selects the goal pose without considering the robot's initial state, resulting in approximately a $2$ ms increase in computation time per step compared to our \textit{re4mpc-PPO} method, which strategically selects suitable intermediate targets. Although \textit{ocs2wb} exhibits fewer total computations during these tests, this reduction is primarily due to its higher failure rate, causing fewer steps overall compared to our variants. Furthermore, the trajectories computed by \textit{ocs2wb} are inherently time-optimal, whereas our method requires additional training to converge to a time-optimal policy while determining sub-targets.

% \begin{figure}[t]
% \centering
% \includegraphics[width=\linewidth]{figures/testing_result_pie.png}
% \caption{Pie charts for the breakdown of the testing results in terms of percentages of success and failure modes.}
% \label{fig:testing_result_pie}
% \end{figure}

%After $100k$ training timesteps, none of the trained models converge to the optimal mobile base trajectory like the greedy baseline method, ocs2wb. Though the average time difference between the baseline and ours is limited to only $2$ action steps, as illustrated in Table~\ref{tab:testing_results}. Learning-based methods often struggle with large action spaces when exploring optimal policies. To address this, either more training time is needed or the reward functions must be better designed to guide the agent toward these optimal policies.
 
% \begin{figure}[t]
% \centering
% \includegraphics[width=\linewidth]{figures/testing_time_arena.png}
% \caption{Average action timesteps per episode.}
% \label{fig:testing_time_arena}
% \end{figure}

\begin{figure}[t]
\centering
\includegraphics[width=0.87\columnwidth]{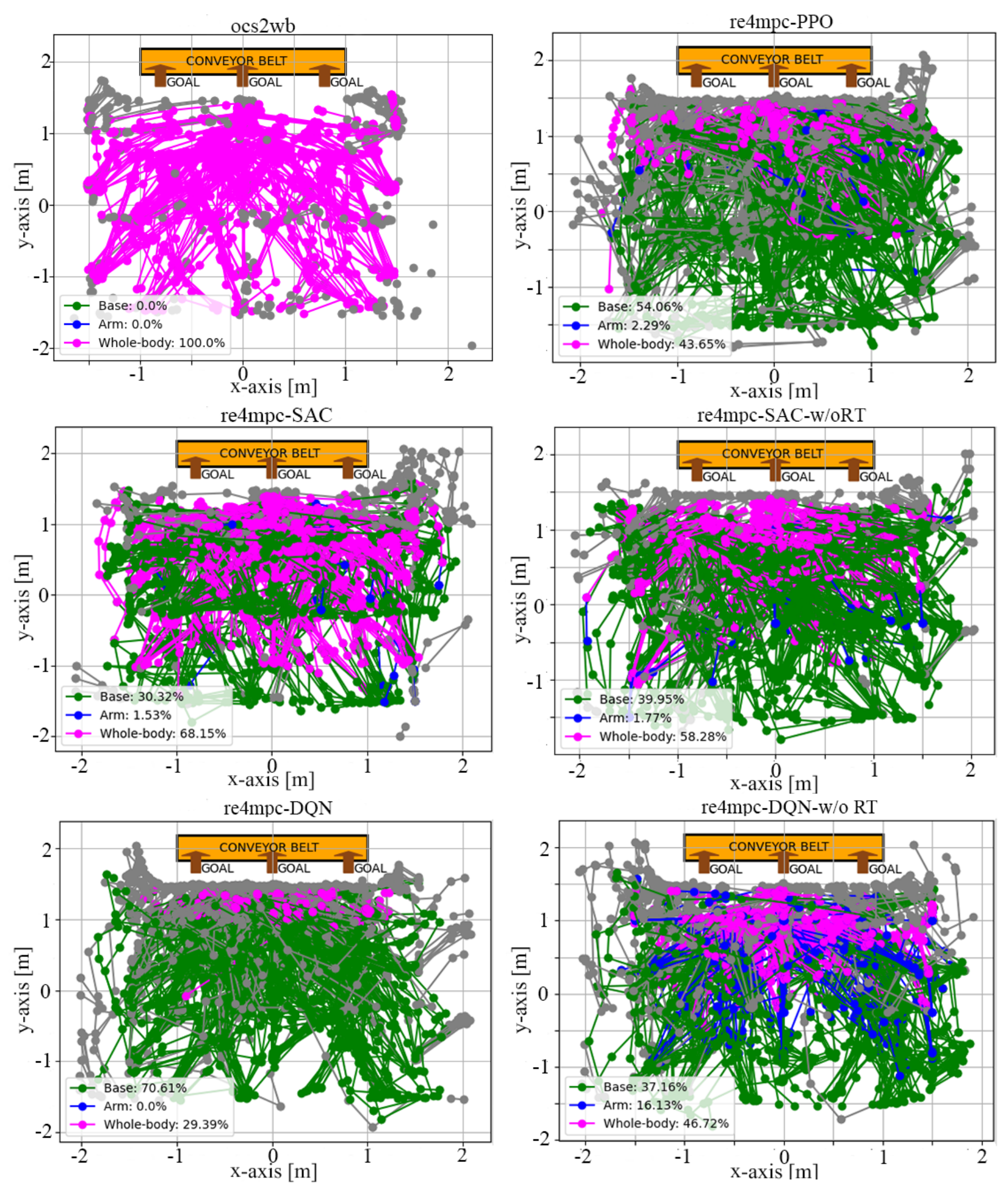}
\caption{Map of robot trajectories shown as connected dots, where each dot represents the robot's position at a time step. Colors indicate the model used: green for base, blue for arm, and magenta for whole-body models. Gray trajectories denote failures to reach the goal. Legend shows each model’s selection frequency across all experiments.}
\label{fig:testing_model_mode}
\vspace{-4.5mm}
\end{figure}

\begin{figure}[t]
\centering
\includegraphics[width=0.87\columnwidth]{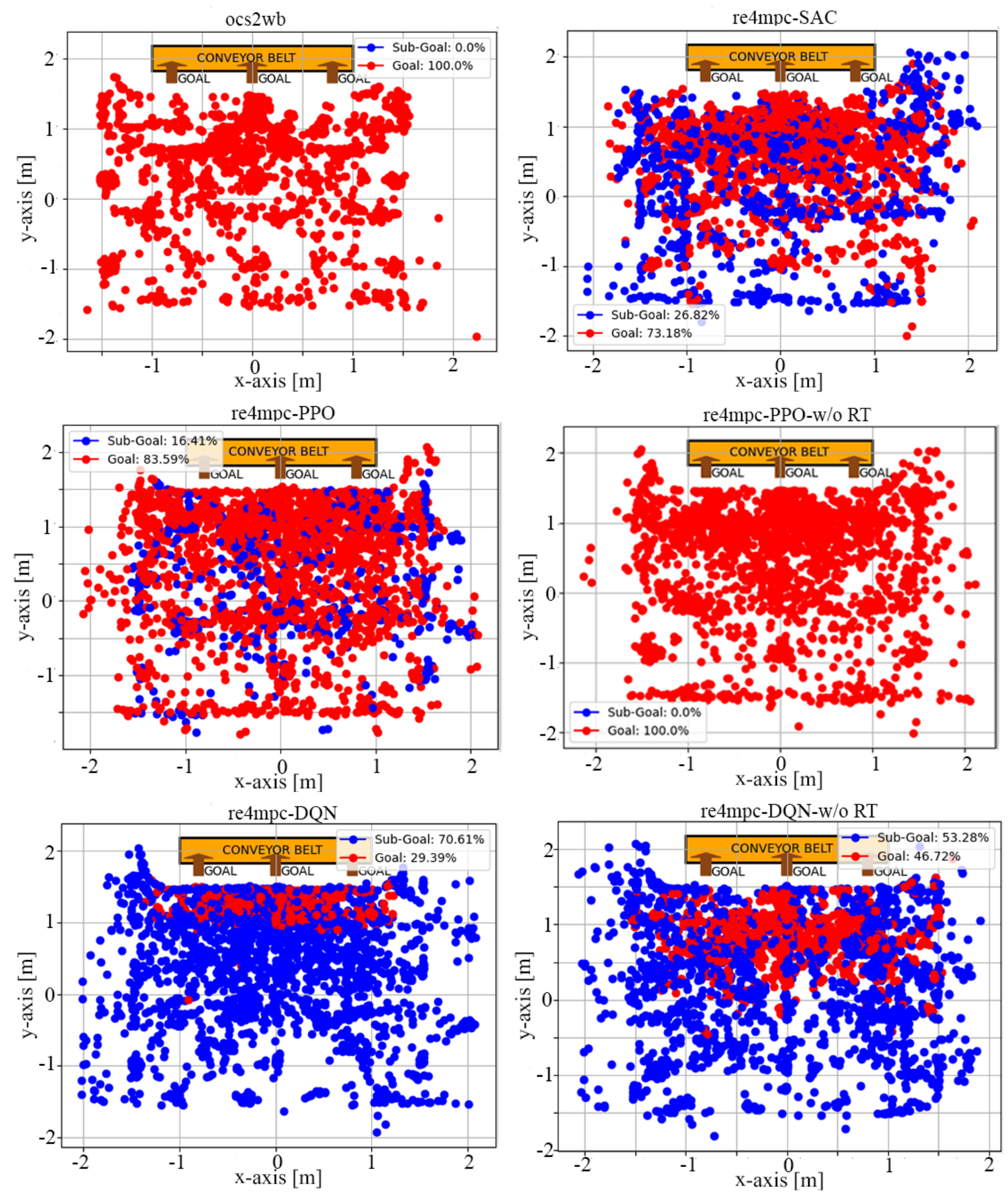}
\caption{Selected target types in the robot’s workspace. Red paths indicate cases where the target was set as the goal, while blue paths show sub-goals chosen by the policy. The legend displays the selection frequency of each target type across all experiments.}
\label{fig:testing_target_type}
\vspace{-4.5mm}
\end{figure}

In addition to success rate, we also explore test set performance based on robot model alterations and target types. For each benchmark method, we plot a 2D map of all generated trajectories, represented as connected dots in Fig.~\ref{fig:testing_model_mode}. Each point indicates the robot's position at that timestep, color-coded to indicate the robot model used to calculate the trajectory. Similarly, Fig.~\ref{fig:testing_target_type} displays chosen \textit{target types}, with red for a goal and blue representing sub-goals. Our findings indicate that Re4MPC significantly reduces whole-body model usage when computing trajectories. For instance, \textit{re4mpc-SAC} uses only base motion for approximately $30\%$ of its successful episodes.

% \begin{table}[t]
% \caption{\nua{MPC Computation time with respect to different robot models}}
% \footnotesize
% \centering
% \setlength\tabcolsep{2.5pt} % Essential for spacing
% \begin{tabular}{c|c|c|c}
% \toprule
% \textbf{} & \textbf{Base} &  \textbf{Arm} & \textbf{Whole-Body} \\ 
% \midrule
% Max [ms] & 10.99 & 17.61 & 48.15 \\ \hline
% Mean [ms] & 4.72 & 3.09 & 5.66 \\ \hline
% Std-dev [ms] & 0.0 & 0.0 & 0.0 \\
% \bottomrule
% \end{tabular}
% \label{tab:mpc_computation_time}
% % \vspace{-4.3mm}
% \end{table}

In our discrete action space formulation, the neural network's output is mapped to a 2D target pose, sampled using the kinematic model of the mobile base. Notably, the discrete action space has an advantage over its continuous counterpart by filtering out unfeasible poses. The difference in action space designs also influences how target type selection is learned. For the continuous action space, the probability of selecting either a goal or a sub-goal is equal, as the decision threshold splits the range in the middle. Conversely, in the discrete case, the probability of selecting the goal is low, as there is only one discrete action per mode associated with setting goals as the target. Discrete Re4MPC variants thus use more base and arm motions while assigning more sub-goal targets than their continuous counterparts.  

Due to the inherent effectiveness of whole-body NMPC, all Re4MPC variants initially imitate this baseline's policy. This pattern, characterized by using the whole-body model and setting the goal as the target, is visually represented by magenta trajectories for the model mode in Fig.~\ref{fig:testing_model_mode} and red for target type in Fig.~\ref{fig:testing_target_type}. As training progresses, the learning agent shifts to policies that are more efficient and feasible according to the reward function. While the success rate for our basic task is not significantly influenced by target type, we hypothesize that employing reduced-order models and setting feasible targets will become crucial when training for more complex tasks. Target reward would then play a pivotal role in training, as evidenced by Fig.~\ref{fig:reward_episodic_v0}, where the learning curves for both PPO and DQN variants ascend earlier than their $R_{target}$-deficient counterparts. Additionally, Fig.~\ref{fig:testing_target_type} suggests that removing the target reward substantially reduces the use of sub-goal targets in PPO and DQN policies.  

As SAC exhibits strong exploration capabilities~\cite{xia2021relmogen}, the SAC variants of Re4MPC often identify optimal policies earlier in their training, making $R_{target}$ less influential on performance. While DQN also thoroughly explores the action space, the probability of selecting an appropriate action for the current state is lower due to our discrete action space design. Therefore, $R_{target}$ encourages the agent toward optimal policies and accelerates learning by about $5k$ episodes.

%===============================================================================

\section{Conclusion}
\label{sec:conclusion}

In this paper, we introduce a novel motion planning pipeline, Re4MPC, that reformulates the NMPC problem via a DRL framework. The system model, constraints, and target information for NMPC are obtained through a decoded action of the RL agent. To integrate NMPC into the DRL framework, we developed a POMDP model by carefully designing the observation and action spaces, as well as the reward functions. We trained the policy network in our proposed pipeline using three different RL algorithms within a physics-based simulation. Evaluation results demonstrated that the policies trained using Re4MPC achieved higher success rates with fewer failures compared to a baseline NMPC method. Moreover, test set results based on model and target type selections made by the RL agent suggest that our approach significantly reduces reliance on whole-body motion planning when computing trajectories. Therefore, Re4MPC converged to a computationally more efficient motion planning policy for the given task. 

\bibliographystyle{IEEEtran}
\bibliography{root}

\end{document}